# KS-Net: Multi-layer network model for determining the rotor type from motor parameters in interior PMSMs

Kıvanç DOĞAN[1] , Ahmet ORHAN [2]

[1 a] Graduate School of Natural and Applied Sciences, Electrical and Electronic Engineering, Firat University, Elazığ, Türkiye
[1 b] Department of Electrical and Electronic Engineering, Faculty of Engineering, Firat University, Elazığ, Türkiye.
[2] Department of Electrical and Electronic Engineering, Faculty of Engineering, Firat University, Elazığ, Türkiye.
[1] kivanc.dogan@firat.edu.tr, [2] aorhan@firat.edu.tr
[1] (ORCID: 0000-0003-4832-1412), [2] (ORCID: 0000-0003-1994-4661)

**Abstract**
The demand for high efficiency and precise control in electric drive systems has led to the widespread adoption of Interior Permanent Magnet Synchronous Motors (IPMSMs). The performance of these motors is significantly influenced by rotor geometry. Traditionally, rotor shape analysis has been conducted using the finite element method (FEM), which involves high computational costs. This study aims to classify the rotor shape (2D type, V type, Nabla type) of IPMSMs using electromagnetic parameters ($L_d, L_q, \varphi_a$) through machine learning-based methods and to demonstrate the applicability of this approach as an alternative to classical methods. In this context, a custom deep learning model, KS-Net, developed by the user, was comparatively evaluated against Cubic SVM, Quadratic SVM, Fine KNN, Cosine KNN, and Fine Tree algorithms. The balanced dataset, consisting of 9,000 samples, was tested using 10-fold cross-validation, and performance metrics such as accuracy, precision, recall, and F1-score were employed. The results indicate that the Cubic SVM and Quadratic SVM algorithms classified all samples flawlessly, achieving 100% accuracy, while the KS-Net model achieved 99.98% accuracy with only two misclassifications, demonstrating competitiveness with classical methods. This study shows that the rotor shape of IPMSMs can be predicted with high accuracy using data-driven approaches, offering a fast and cost-effective alternative to FEM-based analyses. The findings provide a solid foundation for accelerating motor design processes, developing automated rotor identification systems, and enabling data-driven fault diagnosis in engineering applications.

**Keywords:** Machine Learning, IPMSM, Rotor Shape Estimation, Parameter Identification, Motor Design Optimization, Rotor Geometry.

## 1. Introduction

In modern electric drive systems, the demand for high efficiency, compact structure, and precise control has led to the widespread adoption of advanced motor types, particularly Interior Permanent Magnet Synchronous Motors (Interior PMSMs). One of the most critical factors determining the performance of Interior PMSMs is the geometric structure of the rotor. Rotor shape directly affects key parameters such as torque production, efficiency, magnetic saturation, thermal behavior, and fault tolerance. Therefore, predicting the rotor shape based on motor parameters is of great importance for both design optimization and the monitoring and fault diagnosis of existing systems. In recent years, machine learning algorithms have been effectively employed in various applications in the field of electric machines, including parameter estimation, fault diagnosis, and performance analysis. The preference for data-driven approaches over complex physical models shortens the modeling process and reduces the need for expert knowledge. In this context, developing machine learning-based methods to predict rotor shape from Interior PMSM parameters has emerged as an innovative research area in engineering. Rotor shape encompasses numerous physical characteristics, such as magnet placement, the position of magnetic barriers, and the structure of bridges and ribs. These geometric features directly influence many performance parameters, including torque production, efficiency, magnetic saturation behavior, temperature distribution, and fault tolerance. [1]. The successful outcomes achieved in applications such as parameter estimation, fault diagnosis, and torque and position prediction in PMSMs indicate that a similar methodology can be applied to rotor shape prediction. In particular, the flexibility and speed offered by machine learning algorithms in modeling multidimensional and complex relationships provide significant added value to engineering design processes. Furthermore, the decisive role of rotor topology in motor performance encourages the widespread adoption of such prediction and optimization studies in industrial applications [2], [3], [4], [5], [6]. In traditional methods, analyses related to rotor shape are generally performed using the Finite Element Method (FEM), through which various design scenarios are examined. However, FEM analyses are computationally expensive, and since they must be rerun for each new design, they slow down engineering processes and increase costs [7]. At this point, machine learning (ML)-based methods emerge as an innovative paradigm in motor design. ML algorithms have the potential to achieve highly accurate predictions by learning complex, multivariate relationships from data that are difficult to model physically. Indeed, the literature includes numerous studies demonstrating that various PMSM parameters—such as torque, temperature, and rotor position—can be successfully predicted using algorithms like artificial neural networks and support vector machines (SVMs) [8], [9]. Traditionally, the methods used to determine rotor shape and parameters have largely relied on engineering expertise, experience, and extensive simulations. However, these processes are constrained by high computational costs and long development times. In recent years, the application of machine learning algorithms in engineering disciplines has introduced a new paradigm for solving such complex problems. In particular, regression and classification algorithms can learn from large datasets to model and predict the complex relationships between motor parameters and rotor shape [10]. In one study, three different machine learning algorithms—Support Vector Regression (SVR), Random Forest Regressor, and Polynomial Regression—were compared for the prediction of Interior PMSM parameters. Random Forest Regressor demonstrated the highest R² values, indicating that it best explained the variation of the dependent variables in the model [10]. This finding demonstrates that machine learning algorithms offer higher accuracy and generalization capability compared to traditional methods. In the specific context of rotor shape prediction, however, studies directly based on machine learning classification remain limited. Nevertheless, it is well established that a strong causal relationship exists between rotor shape and electromagnetic parameters ($L_d, L_q, \varphi_a$) [11]. Therefore, accurately predicting and analyzing rotor shape is of great importance in motor design and manufacturing. The success of machine learning algorithms used for rotor shape prediction also depends on the quality and diversity of the dataset on which the model is trained. Simulation data obtained through the finite element method, experimental measurements, and parametric design studies provide a rich source of data for training machine learning models [1], [7]. These data enable the model to learn different rotor topologies and parameter combinations. In this study, the prediction of rotor shape (2D type, V type, Nabla type) was targeted using the electromagnetic parameters ($L_d, L_q, \varphi_a$) of IPMSM motors. For this purpose, a custom artificial neural network model, KS-Net, developed by the user, was comparatively evaluated alongside Cubic SVM, Quadratic SVM, Fine Tree, Fine KNN, and Cosine KNN algorithms. The performance of the models was tested using 10-fold cross-validation, and their effectiveness was compared through accuracy rates and confusion matrix. The originality of this study lies in introducing KS-Net as a data-driven approach for rotor shape classification, providing a fast and cost-effective alternative to traditional FEM-based analyses. The test results revealed that the proposed KS-Net algorithm exhibits strong generalization capability and demonstrates competitive, and in some cases superior, performance compared to traditional algorithms. This study demonstrates the applicability of a data-driven approach in IPMSM motor design and provides a foundation for future applications in parametric optimization, monitoring, and fault diagnosis.

---

[1] Corresponding author

## 2. Materials and Methods

The primary objective of this study is to evaluate the performance of machine learning algorithms and a custom-designed artificial neural network model (KS-Net) in rotor shape classification based on the electromagnetic parameters ($L_d, L_q, \varphi_a$) of IPMSM motors. A high-quality and balanced dataset was generated through systematic simulations based on parametric modeling. The dataset comprises three different rotor topologies, each consisting of 3,000 samples:
- 2D type rotor
- V type rotor
- Nabla type rotor

Since all classes contain an equal number of samples, a balanced learning environment was ensured for the classification algorithms. Each sample is characterized by three fundamental parameters ($L_d, L_q, \varphi_a$) representing the electromagnetic behavior of the rotor. These parameters were derived based on the motor's operating conditions and rotor geometry, processed and analyzed in MATLAB. They were used directly as model inputs without requiring an additional feature engineering step.

Within the classification process, the custom-designed wide neural network, KS-Net, was evaluated alongside classical algorithms such as Cubic SVM, Quadratic SVM, Fine KNN, Cosine KNN, and Fine Tree. All algorithms were trained in the MATLAB Classifier Learner interface and tested using the 10-fold cross-validation method.

The model performances were compared using accuracy rate and confusion matrix analyses. Through this comprehensive evaluation approach, the ability of the developed models to correctly distinguish between different rotor topologies was analyzed in detail. The obtained results demonstrate that electromagnetic parameters are highly determinant of IPMSM rotor shape and that high-accuracy classification is achievable through data-driven methods.

## 2.1. Dataset

In this study, a dataset consisting of the electromagnetic parameters of IPMSM motors—namely, the d–q axis inductances $L_d, L_q$, and the static magnetic flux $\varphi_a$ —was utilized. The dataset comprises a total of 9,000 samples corresponding to three different rotor topologies. Each rotor shape contains 3,000 samples, resulting in balanced classes. Table 1 presents the distribution of the dataset.

**Table 1.** Dataset

| No | Class | Number of Data |
|----|-------|----------------|
| 1  | 2D    | 3000           |
| 2  | V     | 3000           |
| 3  | Nabla | 3000           |

The data were obtained from a simulation framework, and the parameters $L_d, L_q, \varphi_a$ were normalized for each sample. The fact that these parameters represent the electromagnetic structure of the motor and are directly dependent on rotor topology has enabled their use in training the classification algorithms in this study.

## 2.2. Feature Selection and Preprocessing

The parameters $L_d, L_q, \varphi_a$ in the dataset were used directly for classification. No dimensionality reduction or filtering was applied, only standard scaling was performed on the input parameters. All algorithms were trained and tested using the same input vectors. This approach was specifically adopted to ensure the comparability of the algorithms.

## 2.3. Classification Algorithms Used

For rotor shape classification, both traditional and custom-developed machine learning algorithms were employed. KS-Net is a custom artificial neural network architecture developed by the user and designed for classification tasks in the MATLAB environment. The model features a deep feedforward neural network structure consisting of three hidden layers.

- The first hidden layer consists of 128 neurons,
- The second layer contains 64 neurons,
- The third layer consists of 32 neurons.

n each hidden layer, the ReLU (Rectified Linear Unit) activation function was used to enhance the nonlinear learning capability. The model was trained for 1000 iterations to ensure stable parameter optimization and convergence of the learning process. With this structure, KS-Net achieved high accuracy in rotor shape prediction based on electromagnetic parameters and successfully distinguished between the classes. The multi-layer perceptron (MLP) architecture of the model is illustrated in Figure 1. [12].

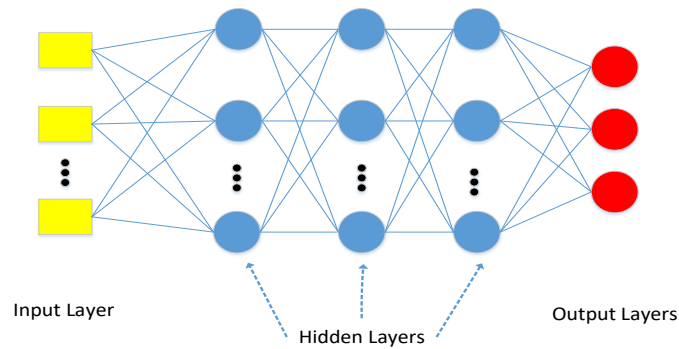

**Figure 2.** 3-layer MLP architecture

The other ML algorithms used for comparison are presented in Table 2.

**Table 2.** Configuration parameters of the proposed classification algoritms

| Classifier | Configuration Parameters | |
|---|---|---|
| Cubic SVM | Kernel function: <br> Kernel scale: <br> Box constraint level: | Cubic <br> Automatic <br> 1 |
| Quadratic SVM | Kernel function: <br> Kernel scale: <br> Box constraint level: | Quadratic <br> Automatic <br> 1 |
| Fine Tree | Maximum number of splits: <br> Split criterion: | 100 <br> Gini's diversity index <br> 100 |
| Fine KNN | Number of neighbors: <br> Distance metric: <br> Distance weight: | 5 <br> Manhattan (Cityblock) <br> Equal |
| Cosine KNN | Number of neighbors: <br> Distance metric: <br> Distance weight: | 10 <br> Cosine <br> Equal |

These models were trained and tested using the MATLAB Classifier Learner Toolbox. A 10-fold cross-validation method was employed for each model, allowing a more reliable assessment of their generalization performance.

## 3. Results and Discussion

In this study, various machine learning algorithms were employed to classify the rotor types of Interior Permanent Magnet Synchronous Motors (IPMSMs). The dataset consists of samples from three different rotor geometries—2D type, V-type, and Nabla-type—where each sample includes the parameters $(L_d, L_q, \varphi_a)$ representing the electromagnetic characteristics of the motor. Six different classification algorithms were evaluated: Cubic SVM, Quadratic SVM, Fine KNN, Cosine KNN, Fine Tree, and the custom deep learning model KS-Net developed by the user. The performance of the models was compared based on accuracy rates and confusion matrices analyses.

The results indicate that the Cubic SVM and Quadratic SVM algorithms classified all test samples without error, achieving 100% accuracy. The proposed KS-Net model achieved 99.98% accuracy with only two misclassifications, reaching a performance level comparable to SVM-based methods. The Fine KNN model misclassified only one Nabla sample and achieved 99.99% accuracy. Cosine KNN produced four misclassifications, resulting in 99.93% accuracy. The Fine Tree model yielded the lowest performance, with a total of 85 misclassifications and 99.06% accuracy, with most errors occurring between the 2D and V classes.

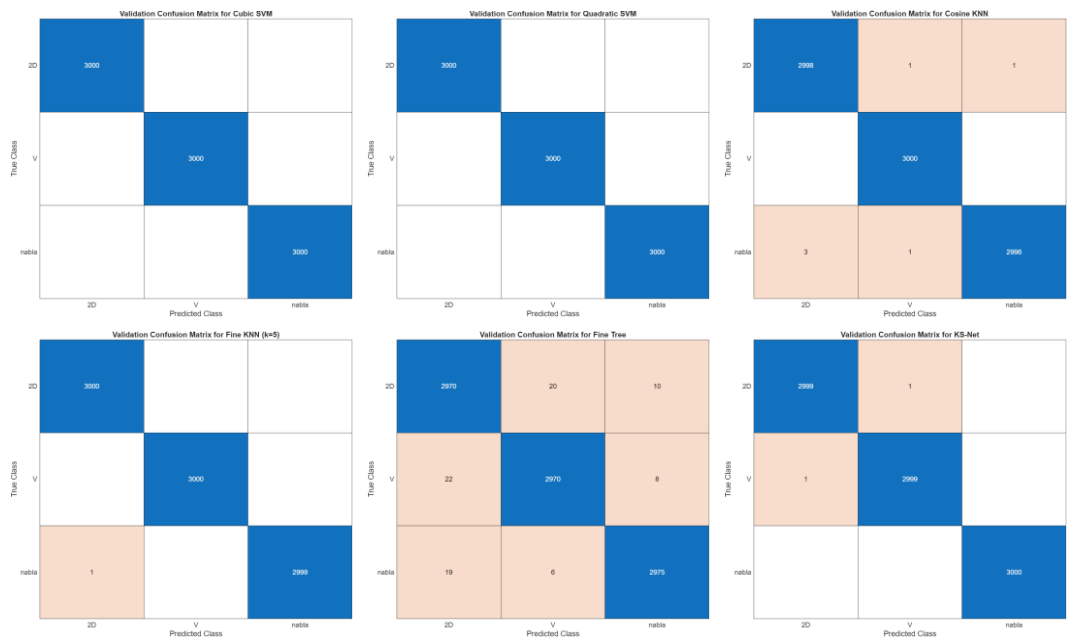

**Figure 2.** The confusion matrices for rotor type prediction correspond, respectively, to Cubic SVM, Quadratic SVM, Cosine KNN, Fine KNN, Fine Tree, and KS-Net.

As shown in Fig. 1, the Cubic SVM and Quadratic SVM algorithms distinguished all classes without error, while KS-Net produced only two misclassifications. The Fine KNN model made a single error in the Nabla class, whereas the Cosine KNN model exhibited four errors, most of which occurred in the Nabla class. In the Fine Tree model, the errors were distributed across classes, with notable confusion particularly between the 2D and V classes. These results demonstrate that the proposed KS-Net model achieves a level of accuracy comparable to classical machine learning algorithms and effectively distinguishes between classes.

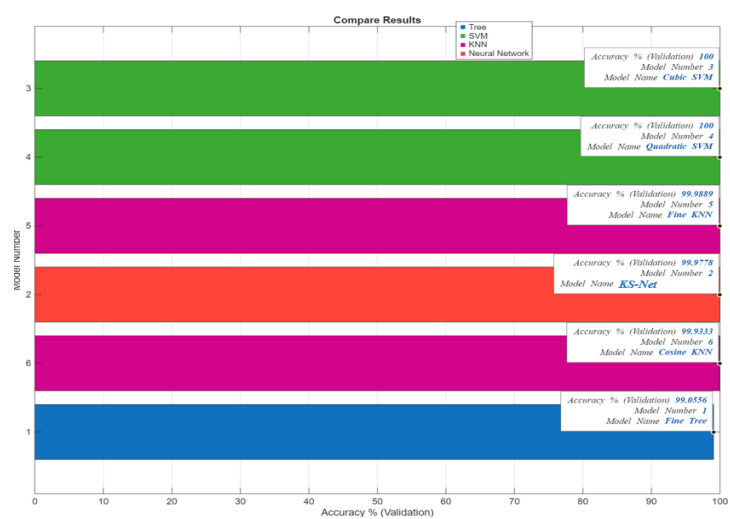

**Figure 3.** Comparison results of classical machine learning algorithms and the KS-Net algorithm

The validation accuracies of all classifiers used in the study are shown in Fig. 2. In the bar chart, the colors represent the algorithm families: blue corresponds to the tree-based model (Fine Tree), green to the SVM-based models (Cubic

SVM and Quadratic SVM), pink to the KNN-based models (Fine KNN and Cosine KNN), and red to the proposed neural network-based model (KS-Net). According to the results, the Cubic SVM and Quadratic SVM algorithms classified all test samples without error, achieving 100% accuracy. This indicates that the dataset can be completely separated using nonlinear SVM kernels and that SVM-based methods are highly effective for this problem. The Fine KNN model achieved 99.99% accuracy with only a single misclassification, exhibiting a performance very close to the SVM-based models. The proposed KS-Net model achieved a similar level of performance with 99.98% accuracy, demonstrating competitiveness with classical methods. Cosine KNN achieved a high accuracy of 99.93%, with a limited number of misclassifications observed. The Fine Tree algorithm yielded the lowest performance with 99.06% accuracy, which can be explained by the inability of decision trees to separate inter-class boundaries as precisely as other algorithms. Overall, the accuracy rates shown in the figure indicate that all models exhibit high success in rotor type classification. However, the SVM-based models demonstrate near-perfect classification capability, while the proposed KS-Net achieves an accuracy comparable to classical methods. These results demonstrate that rotor type classification based on motor parameters can be performed reliably and that the method can be effectively utilized in motor design and automatic rotor type identification. Following this general assessment, a comparative performance table containing the accuracy, precision, recall, F1 score and specificity values of each model is presented in Table 3 to quantitatively illustrate the classification performance of the models.

**Table 3.** Comparative performance metrics of rotor type classification models

| Metrics | Cubic SVM | Quadratic SVM | Fine KNN | Cosine KNN | Fine Tree | KS-Net |
|---|---|---|---|---|---|---|
| Accuracy | 1.00 | 1.00 | 0.9999 | 0.9993 | 0.9906 | 0.9998 |
| Precision | 1.00 | 1.00 | 0.9999 | 0.9993 | 0.9905 | 0.9998 |
| Recall | 1.00 | 1.00 | 0.9999 | 0.9993 | 0.9905 | 0.9998 |
| F1 Score | 1.00 | 1.00 | 0.9998 | 0.9993 | 0.9905 | 0.9998 |
| Specificity | 1.00 | 1.00 | 0.9999 | 0.9996 | 0.9953 | 0.9998 |

The obtained results clearly demonstrate that all algorithms used in this study achieved very high performance in rotor type classification. The Cubic SVM and Quadratic SVM models reached a perfect score of 1.00 (100%) across all metrics, classifying every sample in the dataset without error. This indicates that the dataset is structurally fully separable by SVM-based models. The Fine KNN model achieved performance very close to these two models, yielding values around 0.9999 for all metrics and making only a single misclassification. The Cosine KNN model achieved metric values in the range of 0.9993 to 0.9996, with a total of four misclassifications, demonstrating very high accuracy. The KS-Net model, as the proposed custom deep learning approach, achieved 0.9998 across all metrics, exhibiting performance competitive with classical machine learning algorithms. In contrast, the Fine Tree algorithm showed lower performance compared to the other models, with metrics ranging between 0.9905 and 0.9953. This result can be attributed to decision trees being less precise in defining inter-class boundaries compared to SVM- and KNN-based methods.

In general, the table indicates that the proposed KS-Net model achieved nearly equivalent performance to SVM-based methods and proved to be a strong alternative to classical approaches. The fact that all models achieved accuracy above 99% demonstrates that rotor type classification based on motor parameters can be performed with a high degree of reliability.

**4. Conclusion**

In this study, the rotor shape (2D type, V-type, Nabla-type) of Interior PMSM motors was classified based on their electromagnetic parameters ($L_d, L_q, \varphi_a$) using machine learning algorithms. In this context, the custom-designed neural network model KS-Net was compared with Cubic SVM, Quadratic SVM, Fine KNN, Cosine KNN, and Fine Tree algorithms, and a comprehensive performance analysis was conducted. According to the results, Cubic SVM and Quadratic SVM classified the dataset with 100% accuracy, predicting all samples without error. This indicates that the classes in the dataset are distinctly separable. KS-Net achieved 99.98% accuracy with only two misclassifications, demonstrating performance competitive with the SVM models in terms of overall success. Fine KNN and Cosine KNN also achieved very high accuracies, though some inter-class confusions occurred in certain samples. Fine Tree was the weakest model in terms of inter-class separation, producing more than 60 misclassifications in total. This study presents an effective data-driven approach for rotor shape classification as an alternative to FEM-based analyses. The proposed method is particularly advantageous for accelerating the design process, developing diagnostic systems when rotor geometry cannot be directly measured, and providing a foundation for data-driven fault diagnosis approaches.

In essence, the originality of this study lies in introducing KS-Net as a data-driven method for high-accuracy rotor shape classification, providing a fast and cost-effective alternative to FEM-based analyses and establishing a solid foundation for automated rotor identification and data-driven fault diagnosis in future engineering applications.
In future work, the KS-Net model will be trained on larger and more diverse datasets to enhance its generalization capability, along with hyperparameter optimization and the inclusion of new parametric inputs related to rotor geometry (e.g., rotor volume, air gap). Furthermore, it is planned to move beyond classification toward optimizing rotor design using regression models and developing automated design systems.

**Data available:** https://www.kaggle.com/datasets/uuuuuuuuu/dataset-iron-losses-ipmsms